\documentclass[letterpaper, 10pt]{article}
\usepackage[letterpaper,top=2cm,bottom=2cm,left=2.5cm,right=2.5cm,marginparwidth=1.75cm]{geometry}

\usepackage[T1]{fontenc}

\usepackage{amsmath}
\usepackage{amssymb}
\usepackage{tabularx}
\usepackage{booktabs}
\usepackage{soul, color}
\usepackage[hyphens]{url}
\usepackage[hidelinks]{hyperref}
\hypersetup{breaklinks=true}
\usepackage{graphicx}
\usepackage{caption}
\usepackage{anyfontsize}
\usepackage{subcaption}
\usepackage{cite}
\usepackage{xspace}
\usepackage{comment}
\newcommand{\ie}{\textit{i.e.}, }
\newcommand{\eg}{\textit{e.g.}, }

\newcommand{\meco}{MeCO\xspace}
\newcommand{\doctype}{manuscript\xspace}
\newcommand{\imu}{AHRS\xspace}

\DeclareCaptionFont{xxviii}{\fontsize{9}{9}\selectfont}
\title{\LARGE \bf
Design and Development of the MeCO Open-Source Autonomous Underwater Vehicle}

\author{David Widhalm, Cory Ohnsted, Corey Knutson, \\Demetrious Kutzke, Sakshi Singh, Rishi Mukherjee,\\  Grant Schwidder, Ying-Kun Wu, and Junaed Sattar
\thanks{This work was supported in part by the National Science Foundation Grant \#2220956, and the Science, Mathematics, and Research for Transformation (SMART) Program.
The authors are with the Department of Computer Science \& Engineering, University of Minnesota--Twin Cities, Minneapolis, MN 55455, USA
{\tt\small \{widha008, ohnst034,
knuts983, kutzk015, sing0975, mukhe100, schwi488, wu001210, junaed\}@umn.edu}}%
}
\date{}

\begin{document}

\maketitle
\thispagestyle{empty}
\pagestyle{empty}

\begin{abstract}

We present \meco, the \textit{Me}dium \textit{C}ost \textit{O}pen-source autonomous underwater vehicle (AUV), a versatile autonomous vehicle designed to support research and development in underwater human-robot interaction (UHRI) and marine robotics in general. 
An inexpensive platform to build compared to similarly-capable AUVs, the \meco design and software are released under open-source licenses, making it a cost effective, extensible, and open platform.
It is equipped with UHRI-focused systems, such as front and side facing displays, light-based communication devices, a transducer for acoustic interaction, and stereo vision, in addition to typical AUV sensing and actuation components. 
Additionally, \meco is capable of real-time deep learning inference using the latest edge computing devices, while maintaining low-latency, closed-loop control through high-performance microcontrollers. 
\meco is designed from the ground up for modularity in internal electronics, external payloads, and software architecture, exploiting open-source robotics and containerarization tools. 
We demonstrate the diverse capabilities of \meco through simulated, closed-water, and open-water experiments. 
All resources necessary to build and run \meco, including software and hardware design, have been made publicly available. 
\end{abstract}
\section{Introduction}
\label{sec:introduction}
Autonomous underwater vehicles (AUVs) have seen significant advances, aided by major investments from academia, industry, and government. 
However, most AUVs tend to be expensive, complex, and proprietary systems, requiring significant effort to deploy and use \cite{mcfarlane1997auv,westwood2001global,offshoremagazine2023, bae2023survey}. 
Cost effectiveness, portability, extensibility, openness, and community support are extremely desirable attributes for AUVs, as these help lower costs and improve repairability, lower the barrier of entry, ease deployment, enable future expansions with updated systems, and importantly, help sustain replicable, field-validated research.
Additionally, there are many use cases that merit human-robot teaming, such as underwater archaeology, crime scene investigation, explosive ordnance disposal, ship hull and pipeline inspections \cite{liang2005experiment, rymansaib2023prototype,sub2012conceptual,hong2019water,rumson2021application}. 
AUVs could provide operational support by performing the tasks that are dirty, dull, or dangerous for humans \cite{takayama2008beyond}. But having strong underwater human-robot interaction (UHRI) capabilities, would enable collaboration between divers and AUVs. 
AUVs could thus exploit human skills and knowledge, effectively becoming an extension of the domain experts, instead of replacing long-established human expertise outright. 
Such \textit{in situ} mission configuration thus requires UHRI technologies to ensure natural, safe, and robust operations given the proximity between the humans and robots.

\begin{figure}
\centering
    \includegraphics[width=0.75\columnwidth]{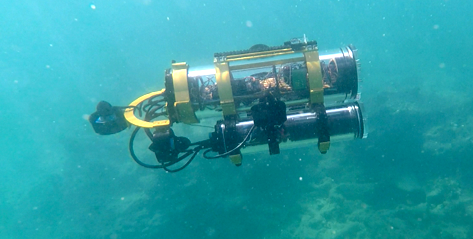}
    \caption{The \meco AUV operating during an open-water field trial in the Caribbean Sea off the coast of Barbados.}
    \label{fig:glamor}
\end{figure}

In this \doctype, we introduce the \meco AUV (Fig.~\ref{fig:glamor}), designed to be an open-source, modular, inexpensive, extensible, and person-portable companion-AUV (co-AUV). 
It features a unique vehicle body that supports tethered (\ie remotely operated) and untethered (\ie autonomous) missions, while featuring a suite of underwater, human-robot interfaces.
\meco parts cost approximately $10$K USD, which is over an order of magnitude less than AUVs with comparable features (\eg \cite{aquacost}).
The ``tri-tube'' design conveniently modularizes the vehicle subsystems to distribute power, sensing, data processing, communication, and multimodal feedback to support a variety of mission contexts, and remain extensible.
While designed to be fully autonomous, \meco is also capable of direct, bidirectional interaction with teams of divers in multi-UHRI missions. 

This \doctype describes our contributions towards the physical, electrical, and software designs, as well as behavioral capabilities of the MeCO platform in creating a UHRI-capable, open-source AUV. Our primary contributions are 
\begin{itemize}
    \item an open-source AUV for general marine robotics and UHRI research and applications, 
    \item the introduction of a simulator for virtual interaction with \meco for prototyping behaviors, and 
    \item the presentation of rigorous open-water field testing of a sample of capabilities provided by \meco.
\end{itemize}
The documentation to build, operate, and extend \meco is available at: \url{https://github.com/MeCO-AUV}.

\section{Related Work}
\label{sec:related_works}
Given the vast body of literature on AUVs, we limit our discussion to small sized vehicles that are not built for long duration autonomy operations (\eg gliders \cite{alvarez2009folaga,wang2022development} or large diameter vehicles \cite{furlong2012autosub,Anduril,orcauuv,Petrel}).

\meco is inspired by and extends the capability of the LoCO AUV \cite{LoCo}. Unlike LoCO, which is limited to three degrees of freedom (DoF) motion control, \meco has non-holonomic 6-DoF control, permitting fine-tuned movement in small operational spaces. This is inspired in part by the CUREE robot designed by the Woods Hole Oceanographic Institute (WHOI) \cite{girdhar2023curee}, which features six DoF control with six thrusters. By comparison, \meco achieves six DoF control with five thrusters. 
\meco has a stereo vision system, along with upgraded sensing and actuation modules compared to the LoCO AUV. The Modularis AUV \cite{modularis} also extends the capability of LoCO; however, it does not provide the onboard processing power or positioning accuracy that \meco needs to perform onboard deep learning inference tasks and complex navigation maneuvers within the diver's operational space. Modularis AUV leverages the Nvidia Jetson Nano computer for computational tasks, but certain simulations of runtime scenarios proved too computationally intensive for the Modularis AUV's Jetson Nano. Anticipating significant real-time computational requirements, we configured \meco with an Nvidia Jetson Orin NX that provides additional computing power over the Jetson Nano \cite{nvidiacomp}.

Some recently developed AUVs offer users highly specialized functionality. 
The HippoCampus AUV \cite{hippox}, for example, performs well in confined environments, such as nuclear waste ponds, given its small form factor ($0.30\times0.05\times0.05$~m) but does not expose UHRI capabilities that could serve commercial divers working in these hazardous conditions \cite{vallance2004radiation,nuclearwastediver}. 
The Cuttlefish AUV \cite{cuttlefish}, by contrast, is larger than HippoCampus ($2.8\times2.0\times0.8$~m), making confined space navigation more challenging. 
However, its larger size and controllability allow it to maintain arbitrary orientations in the water column and is designed specifically for manipulation tasks like underwater construction or pick-and-place operations. 
However, it does not provide on-board R2H or H2R communication modalities, making it unsuitable for prototyping UHRI algorithms. 

Chen et al. \cite{UAH} introduced a novel-shaped platform for studying helicopter-style maneuverability in an AUV. 
This is a stark contrast to the traditional streamlined AUV shape often adopted by most designs, given their hydrodynamic advantage. 
These autonomous underwater helicopters (AUHs) minimize turning radii, greatly improving maneuverability, yet they are not ideal companion AUVs, given their lack of UHRI interfaces.

Additionally, many AUVs that are available for purchase such as the REMUS family of vehicles and the Aqua AUV (see \cite{allen1997remus} and \cite{dudek2007aqua}, for example) are often more expensive than \meco ($1$M USD for top-end models and $150$K USD, respectively, compared to $10$K USD for \meco) and do not benefit from community developed contributions and stability that come with an open-source platform.

The Boxfish AUV \cite{boxfish} is another commercially available system for general purpose marine robotics applications. With an eight thruster configuration, it can achieve sophisticated maneuvering. It is marketed as a general purpose AUV. While there are many commercial AUV options that are some combination of small sized, extensible, and maneuverable, there is no single vehicle that fills the gap between purely off-the-shelf capability and open-source extensibility that encourages research in the traditionally under-served UHRI research space. 
\meco fills the gap by providing a platform that meets maneuverability expectations, open-source UHRI interfaces, and field-tested capabilities for broad-spectrum UHRI development and general marine robotics.

\section{Physical Characteristics}
\label{sec:physical_design}

\begin{figure}
    \centering
    \includegraphics[width=0.75\columnwidth]{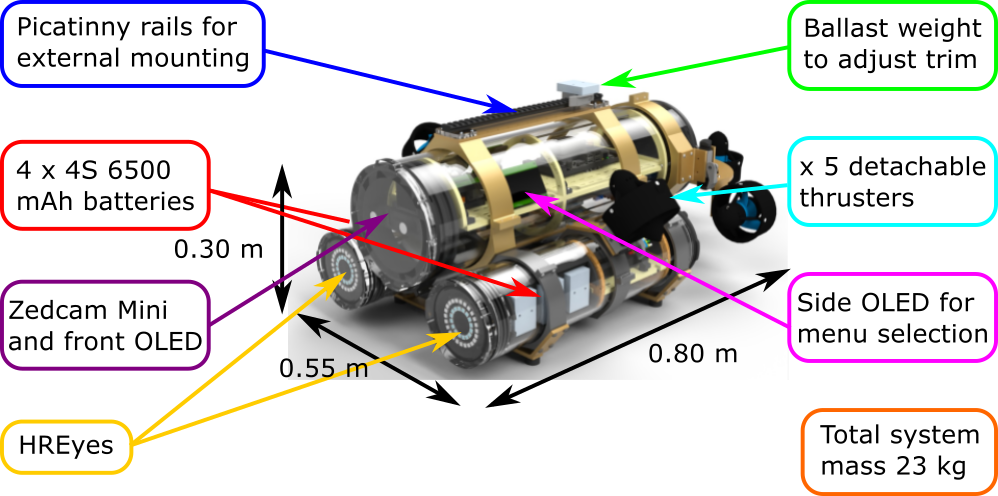}
    \caption{\meco CAD model highlighting dimensions, power, sensing, and propulsion features.}
    \label{fig:label_models}
\end{figure}

The design of \meco facilitates the modular, reconfigurable, and user-oriented nature of its philosophy. Fig.~\ref{fig:label_models} shows a high-level overview of \meco's features.
It consists of three acrylic tubes and an external anodized aluminum structure which holds the tubes in place and also provides external mounting points for additional hardware. 
The center tube is $80$ cm long and has a diameter of $16.5$ cm. 
The side tubes are $38$ cm long and have a diameter of $11.4$ cm.

\meco has a buoyant volume of $23.6$ liters, which leads to a neutrally buoyant weight of $23.6$ kg in freshwater, and $24.2$ kg in salt water. 
With no ballast, \meco weighs $20.9$ kg and has ample room inside each tube for placement of ballast. 
This allows for reconfiguring of the center of gravity to suit a particular mission or control style. 
\meco can be made vertically stable by shifting the internal ballast lower, but is currently configured to have the center of gravity close to the center of buoyancy to allow it to hold arbitrary orientations.

The clear acrylic housings provide a number of advantages. Cameras can be placed virtually anywhere in or on the robot for vision, indicator lights can be easily seen to check on various systems, screens can be housed inside the robot to provide detailed information, and any damage or leaks can be seen so they can be handled quickly. Currently a front facing stereo camera and system indicator lights are placed in the center tube.

The main components of the aluminum structure are clamps which hold the tubes in place via friction. 
While the current layout of \meco shown here has the three tubes lined up at the forward end, the side tubes can be moved in relation to the center tube for different configurations if desired simply by sliding them within their clamps.

The structure also provides mounting points for external hardware via Picatinny rails. 
These rails allow for items to be quickly clamped on to the robot in various locations. 
These clamps can be moved easily in the field for quick reconfigurations of payload or ballast. 
Four of the thrusters are also mounted via Picatinny clamps and they can be similarly adjusted in the field to adjust motion dynamics.

\begin{figure}[t]
    \centering
    \includegraphics[width=0.6\columnwidth]{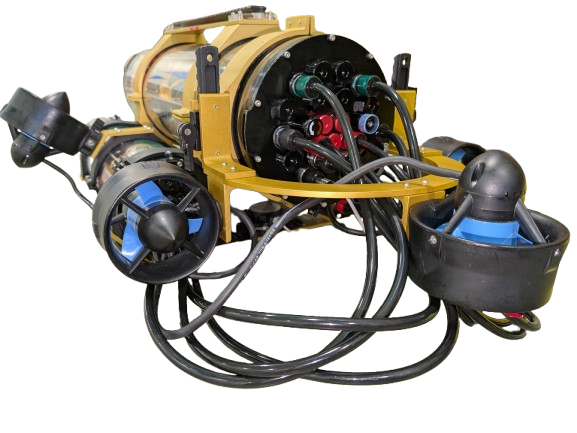}
    \caption{Rear view perspective of \meco, showing cable routing to distribute power, PWM signals, Ethernet, and common ground between all three tubes.}
    \label{fig:power_routing}
\end{figure}

\meco currently has five thrusters arranged around the robot. 
Two are aligned with the surge axis, one with the heave axis, and two are oriented $45$ degrees between the heave and sway axes. 
Depending on the center of gravity, each thruster also provides a rotational moment to the robot which affects the dynamics. 
As previously discussed, with the exception of the heave thruster, each can be adjusted slightly in the field to account for some changes in dynamics. 
\section{Electronics}
\label{sec:electronics}

\meco distributes power between its three tubes by individually powering each side tube, and routes battery power from the side tubes to the center tube.
Fig.~\ref{fig:power_routing} shows the cabling and routing configuration of the power system on the actual vehicle. Power and signals can be passed between each tube through wires connecting the tubes protruding from the aft bulkheads.
A common ground reference between each tube keeps communication signals consistent by reducing noise and minimizing signal interference \cite{gupta2011_electricalground}.
\meco's power distribution system also ensures that a single side tube can power the entire vehicle, effectively increasing reliability by virtue of power system redundancy. 
The main computing system is an Nvidia Jetson Orin NX\texttrademark{} \cite{nvidiacomp} housed in the center tube. Low-level functions are executed via the Attitude Heading Reference System (\imu) which is in the center tube located near the expected center of gravity of \meco, Teensy\texttrademark{} 4.1  microcontrollers \cite{teensy} in the side tube for motor controls and in the center tube for menu display.

A high-speed solid-state drive (SSD) mounted in the center tube provides space for mission data storage. 
Communication between the microcontrollers and the main computer is handled via Ethernet, and an Ethernet switch is mounted in the center tube which also provides the port for the detachable Fathom\texttrademark{} tether \cite{bluerobo} during tethered operations.
The tether is an Ethernet-only connection for passing communication between a ground station and the robot, and does not allow external power.

Each tube has a separate $5$~V power bus referenced to the common ground for ancillary electronics. Currently, four Lithium-Polymer (LiPo) batteries housed in the side tubes provide a total capacity of $385$ watt-hours. 
At idle, which we consider to be the lowest power consumption state, with no thrusters running or no heavy computational loads, \meco draws 1.9 amps of current, which equates to an idle running time of 13.5 hours. 
With all five thrusters running and the computer running a heavy computational load (such as deep learning inference), which is the highest power consumption state, the total current draw is approximately 115 amps, which is a run time of 0.2 hours. 
This is an unlikely load as all five thrusters generally do not run simultaneously at full throttle. 
The expected runtime is between these extremes, but in field operations lasting multiple hours, battery changes have for \meco have not been necessary.

\section{Underwater Human-Robot Interaction Capabilities}
\label{sec:uhri_cap}
\begin{figure}[t]
    \centering
    \begin{subfigure}{0.39\columnwidth}
        \centering
        \includegraphics[width=\textwidth]{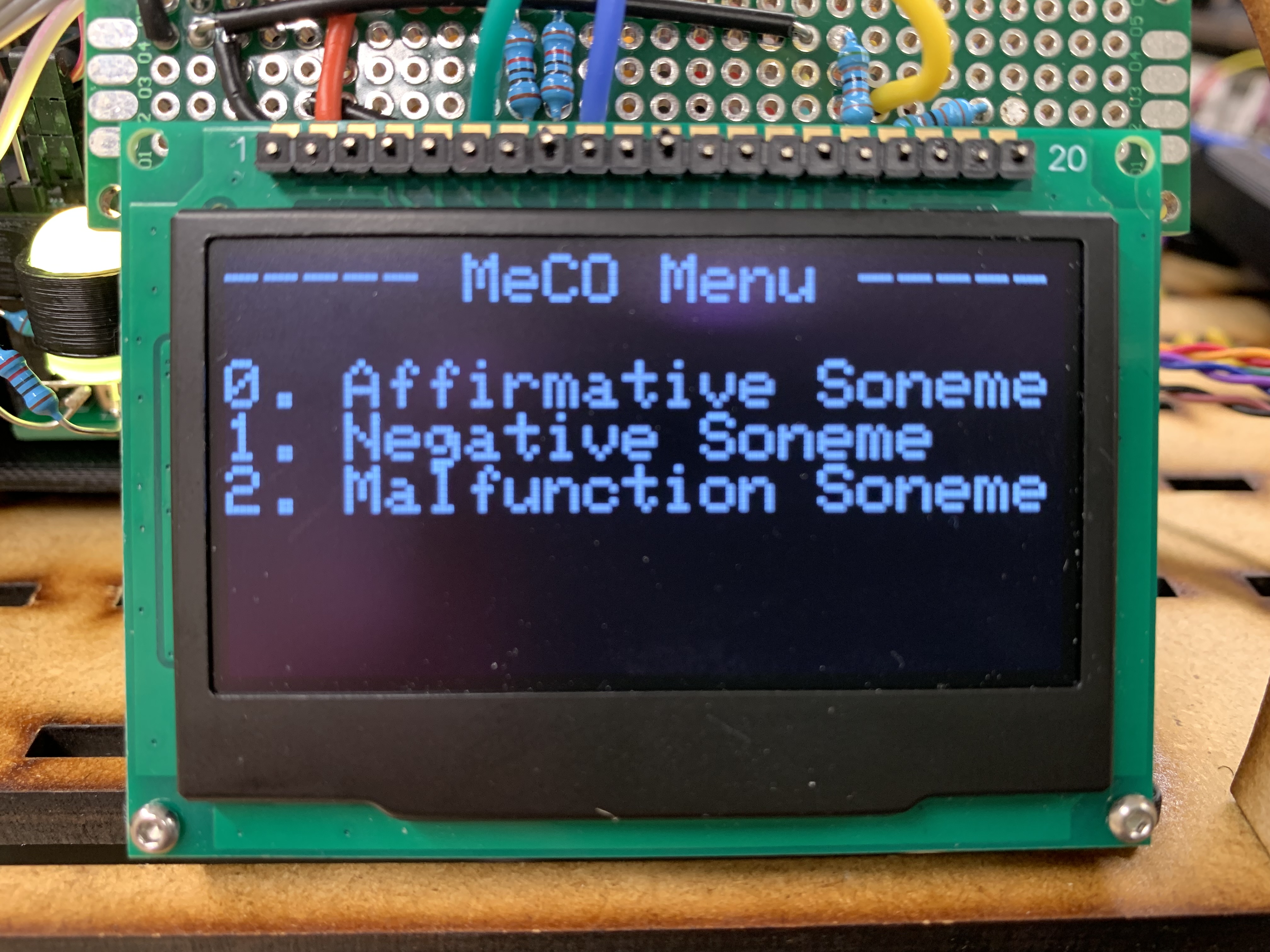}
        \caption{}
        \label{fig:oledside}
    \end{subfigure}    
    \begin{subfigure}{0.3\columnwidth}
        \centering
        \includegraphics[width=\textwidth]{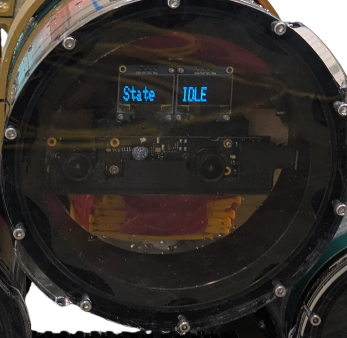}
        \caption{}
        \label{oledfront}
    \end{subfigure}
    \caption{OLED displays on MeCO. (a) A customizable experiment menu is displayed on the side OLED. (b) The front OLED display showing ``State IDLE," during one such experiment.}\label{fig:oled}
\end{figure}

To facilitate UHRI, \meco includes three R2H communication systems: 
\begin{enumerate}
    \item OLED displays: Three organic light-emitting diode (OLED) displays serve as one of the visual communication tools. These are housed in the center tube. One on the side shows the experiment menu to the diver (Fig.~\ref{fig:oledside}) and two in the front shows the status of the robot (Fig.~\ref{oledfront}).
    \item HREyes: HREyes~\cite{fulton2022hreyes} are a light-based visual communication system used on MeCO. These consist of one $16$ and one $24$ RGB LED Neopixel ring concentrically aligned to display color-coded messages, or mimic ocular cues. An HREye module is attached to the front cap of each side tube. In Section \ref{sec:experiments}, we demonstrate HREyes functionality as part of an object detection task (Fig.~\ref{fig:trash_hreye}).
    \item SIREN: SIREN~\cite{fulton_siren_2023} is an audio-based communication system on MeCO that utilizes two distinct modalities: synthesized text-to-speech (TTS) and tonal indicators. This system is mounted on the lower Picatinny rail under the center tube.
\end{enumerate}

MeCO also includes two H2R interfaces that permit interaction via visually provided information:
\begin{enumerate}
    \item Gesture-based control: Employs deep learning methods for robotic detection and classification of a diver's gesture underwater. 
    \item Tag-based control: Using Augmented Reality (AR) tags, the diver can interactively control the robot as an alternative method to gesture-based control. 
\end{enumerate}

\noindent These methods are explained further in Section~\ref{sec:experiments}.

\section{Software}
\label{sec:software}

\begin{figure*}[t]
    \centering
    \includegraphics[width=1.0\textwidth]{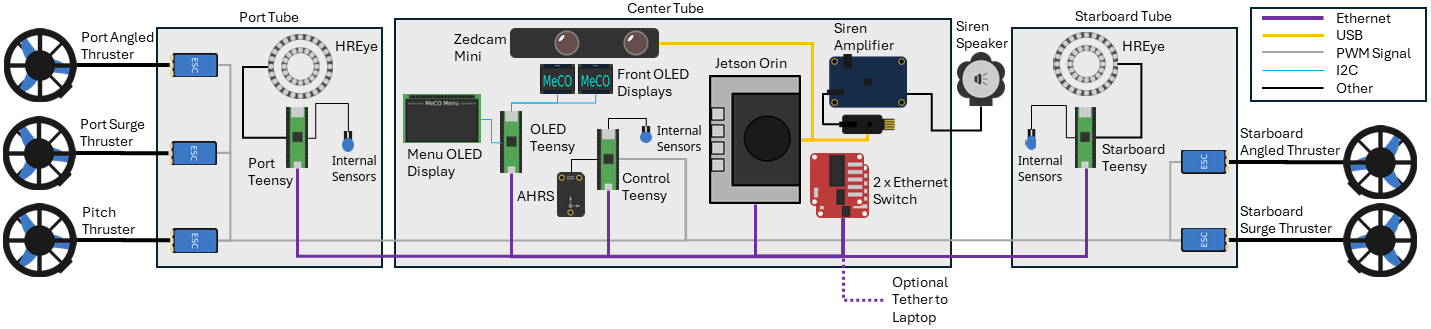}
    \caption{\meco subsystems diagram showing connections between side and center tube components.}
    \label{fig:systems_diagram}
\end{figure*}

With extensibility being a core design principle, we require \meco's software system to execute multiple concurrent processes without interfering with core functionalities of the robot or with each other. The MeCO software system is thus built using tools such as Robot Operating System (ROS 2) \cite{ros2}, micro-ROS \cite{belsare_micro-ros_2023}, and Docker \cite{merkel2014docker} to satisfy this design philosophy and provide a robust software foundation.

The software stack on \meco is divided between two platforms: the Nvidia Jetson Orin NX and the Teensy $4.1$ microcontrollers. 
Code that will benefit from  CUDA acceleration, or code that requires any significant amount of memory or disk space is written as ROS 2 nodes, containerized, and run on the Nvidia Jetson Orin NX. 
Docker serves as a containerization tool, allowing the isolation of core robot functionality. 
This allows the user to build and test new behavior on top of the base system without affecting the core functions. 
Moreover, for perception-based systems that rely on CUDA acceleration and varying versions of PyTorch, TensorFlow, and other deep learning frameworks, Docker provides a ``payload autonomy" model by reducing integration errors and decreasing development timelines \cite{mabry_maritime_2016}.

Timing-sensitive code subject to soft real-time constraints, such as control loops or low-latency data publishing, runs on the Teensy 4.1 microcontrollers. 
Micro-ROS facilitates the communication with the ROS 2 environment on the Jetson Orin. 
This communication is handled via the Data Distribution Service - eXtremely Resouce Constrained Environment protocol (DDS-XRCE) over Ethernet. 

All sensors and actuators, except for the camera, interface directly with Teensy microcontrollers to publish and subscribe to data in ROS 2 (see Fig.~\ref{fig:systems_diagram}). The camera is the only component directly connected to the Jetson Orin. The use of Docker containers and ROS 2 topics offer a modular framework for software development. This architecture simplifies the process of implementing new features incrementally on a reliable base platform.
\section{Simulation}
\label{sec:simulation}

It is well-known that simulation systems are beneficial for underwater robotics research and applications, because field testing is high-cost, high-risk, and logistically challenging \cite{prats2012open,potokar2022holoocean}. 
We developed a high-fidelity 3D simulator for \meco that is unique compared to existing systems \cite{von2022open, zhang2022dave}, for three primary reasons: our simulator enables complex simulation and mission reconfiguration \textit{in situ}, system errors and measurement uncertainties are decoupled, and \meco's UHRI capabilities are modeled. 
The simulation is built using Unity $2022$\cite{Unity} and mimics the communication data flow of \meco. It uses ROS 2 for communication with the same hardware control that MeCO uses while still supporting keyboard inputs. This allows it to be run using the same control hardware as \meco or as a standalone simulation. 
We will release this simulator with \meco's source code.

\textbf{Dynamic reconfiguration}. As a modular platform, we expect \meco's configuration of both hardware and software to change rapidly. 
Many AUV simulators use static meshes and physical parameters as the base for their agents. 
Our simulator permits \textit{in situ} reconfiguration of the vehicle's buoyancy, structure, thrusters, and equipment---even during runtime.
This enables dynamic reconfiguration to test phenomena such as the effect of the loss of a thruster, for example, since the simulation does not assume any specific vehicle configuration. 
This modeling paradigm also enables vehicle-surface interactions such as the effect of surface waves action. 
We employed this simulator during the \meco development cycle to test various center of gravity and thruster configurations.

\textbf{Uncertainty modeling}. We incorporate hardware system errors and measurement uncertainties as independent variables to give the end user the ability to probe these variables as required without coupling. As an example, the \imu is given an actual location in simulation space but has a separate simulation parameter for storing the measured installation location and orientation so that misalignment can be modeled and tested for.

\begin{figure}
\centering
    \begin{subfigure}[t]{0.35\columnwidth}
    \includegraphics[width=1.0\textwidth]{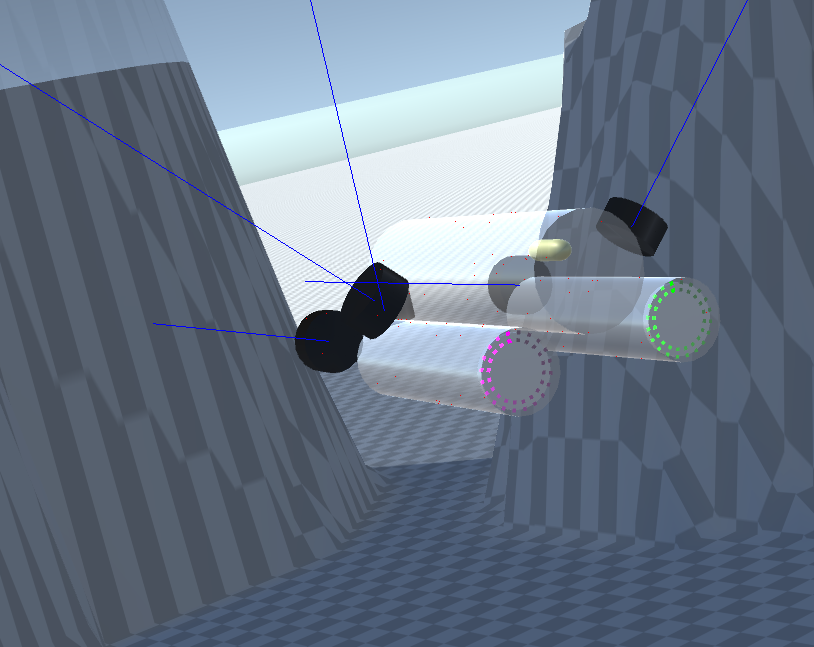}
    \caption{HREyes simulation.}
    \label{fig:meco_sim}
    \end{subfigure}
    \begin{subfigure}[t]{0.35\columnwidth}
    \includegraphics[width=1.0\textwidth]{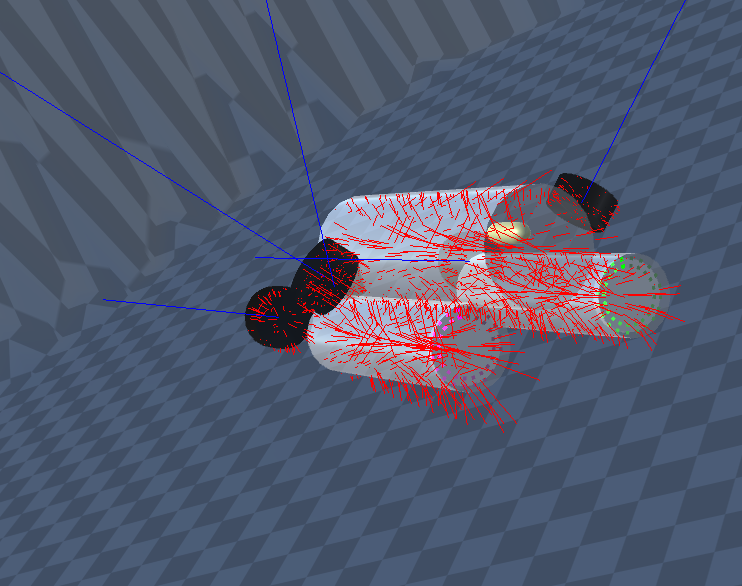}
    \caption{Fluid simulation.}
    \label{fig:meco_drag}
    \end{subfigure}
    \label{fig:meco_simulation_parent}
    \caption{\meco AUV in Unity simulation. Blue lines show thrust directions. Yellow object shows calculated position and orientation from the \imu. Red lines in Fig.~\ref{fig:meco_drag} show draglines from a simplified fluid simulation.}
\end{figure}

\textbf{UHRI capability modeling}. \meco exposes five primary UHRI interfaces for end users to prototype interaction algorithms. The simulation provides virtualization of these capabilities to test UHRI specific algorithms before field deployments. For example, Fig.~\ref{fig:meco_sim} shows the HREyes illuminated during a simulation run. These capabilities are exposed using the same software interfaces as the actual vehicle.

\section{Experiments}
\label{sec:experiments}

We conducted experiments with \meco in three primary environments: closed-water swimming facilities; freshwater lakes of the upper Midwest region, USA; and seawater off the coast of Barbados. To test full-scale functionality, we conducted experiments in these environments in both tethered and untethered operational modes in the freshwater lake environment. We used tetherless UHRI capabilities in the seawater environment for ease of deployment. These experiments demonstrate \meco's versatility as a reconfigurable platform with base capabilities, while also incorporating on-going research. This is for two reasons: to expand \meco's off-the-shelf capabilities and also to gauge the level of convenience and ease with which a user can integrate novel code within the \meco software ecosystem.

These experiments focus on different aspects of \meco's capabilities and showcase how they can be used for a variety of research and application agendas.

\begin{figure}[t]
    \centering
    \begin{subfigure}[t]{0.46\linewidth}
    \includegraphics[width=1.0\linewidth]{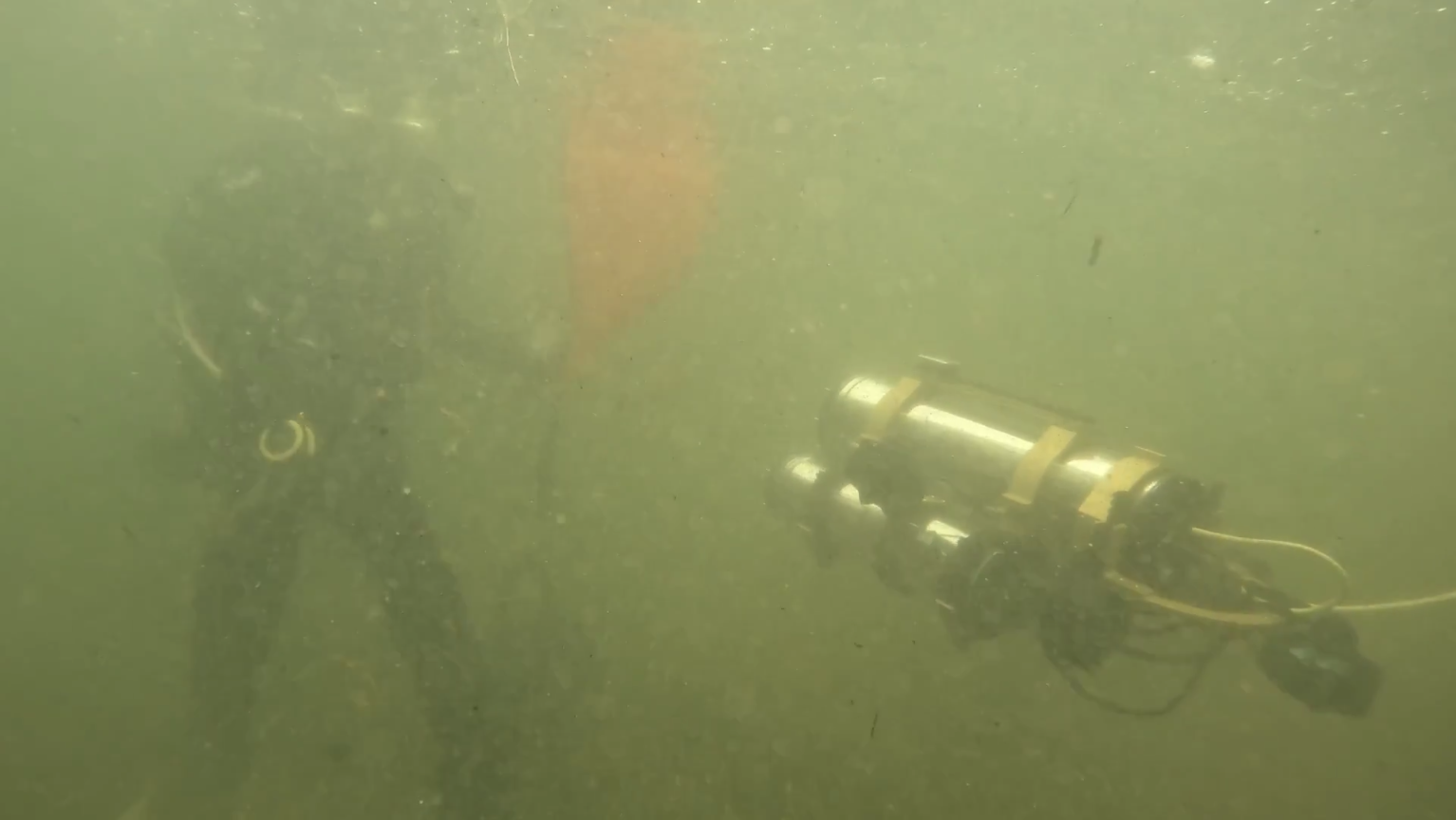}
    \phantomcaption{}
    \label{fig:mav_lake}
    \end{subfigure}    
    \begin{subfigure}[t]{0.48\linewidth}
    \includegraphics[width=1.0\linewidth]{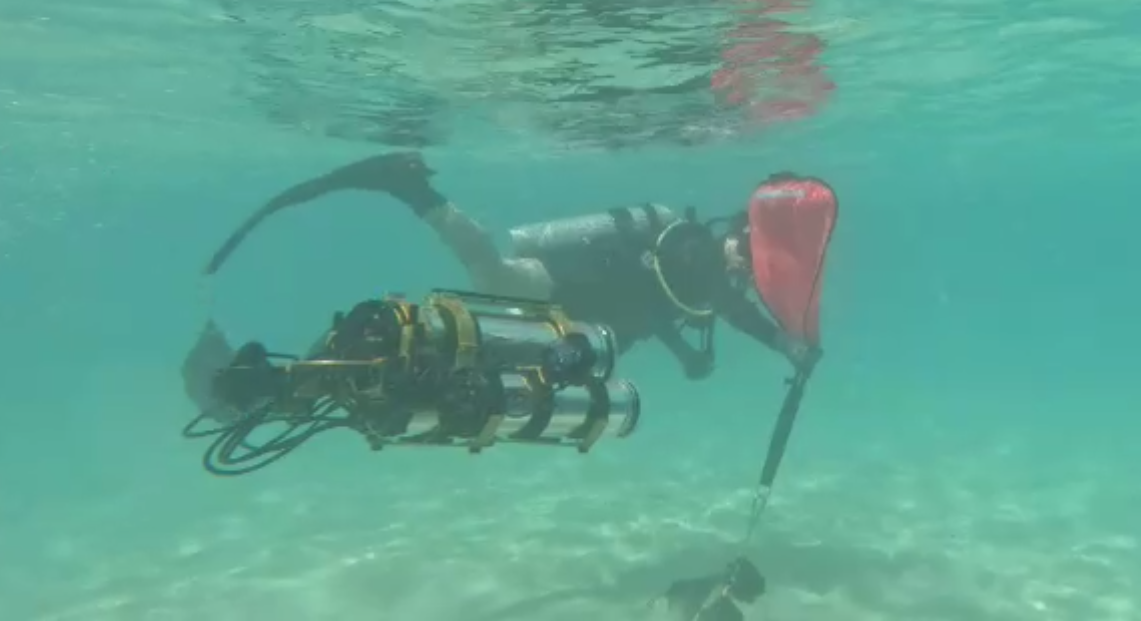}
    \phantomcaption{}
    \label{fig:mav_ocean}
    \end{subfigure}
    \label{fig:mav_comp_parent}
    \caption{\meco AUV running MAV code in both freshwater lakes (top) and saltwater environments (bottom) with different centers of gravity and ballast conditions. \meco's reconfigurability permits fast adjustment and fine tuning for different ballast requirements in various water environments.}
\end{figure}

\noindent \textbf{Movement.} A key design feature of the \meco platform is maneuverability and control with its five thruster configuration. Maintaining camera position and orientation during close-range communication is critical for vision-based applications such as UHRI. Control of \meco is handled through a software controller which can act as a fly-by-wire system, or as a fully autonomous controller. These modes are configured such that changing thruster position or center of gravity will have minimal impact on the control scheme. By using configuration files for thruster parameters and physical attributes \eg center of gravity, the controller can simply use the updated parameters to calculate proper motor commands without further user involvement. This was successfully demonstrated throughout a number of trials. First, \meco had a center of gravity positioned low to keep it stable in the roll axis. A number of experiments were run including the MAV experiments (described below) in a freshwater lake. Second, the center of gravity was raised and moved forward so that it coincided with the center of buoyancy while re-ballasting \meco for saltwater. The MAV experiment was then rerun in a saltwater environment in the Caribbean Sea off the coast of Barbados with minimal effort by changing the parameter file to reflect the new center of gravity. The controller functioned equally well in both cases with \meco following the target successfully as seen in Fig.~\ref{fig:mav_lake} and~\ref{fig:mav_ocean}. 

\begin{figure}
    \centering
    \begin{subfigure}{0.4\columnwidth}
        \centering
        \includegraphics[width=\textwidth]{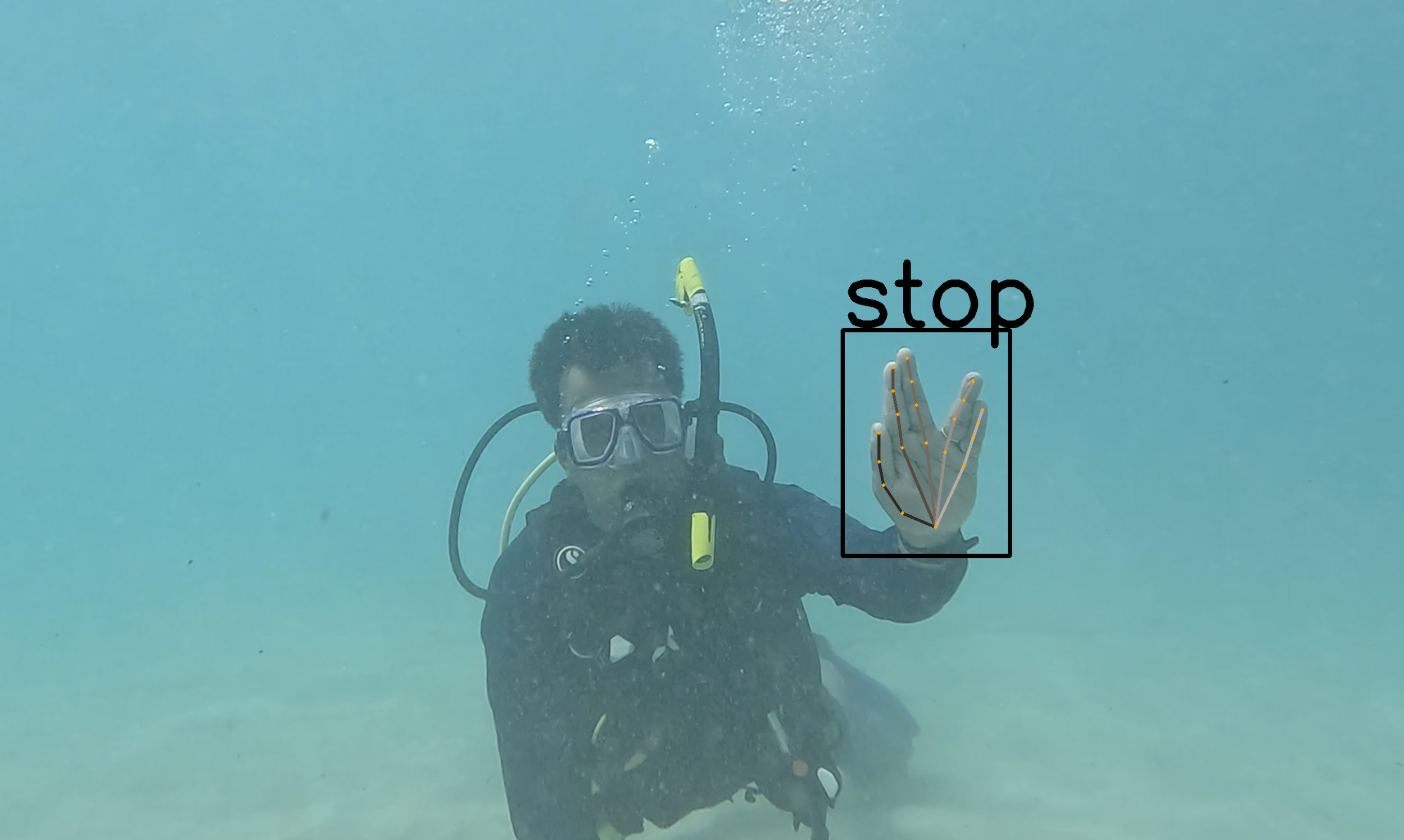}
        \caption{}
        \label{fig:gesture}
    \end{subfigure}    
    \begin{subfigure}{0.4\columnwidth}
        \centering
        \includegraphics[width=\textwidth]{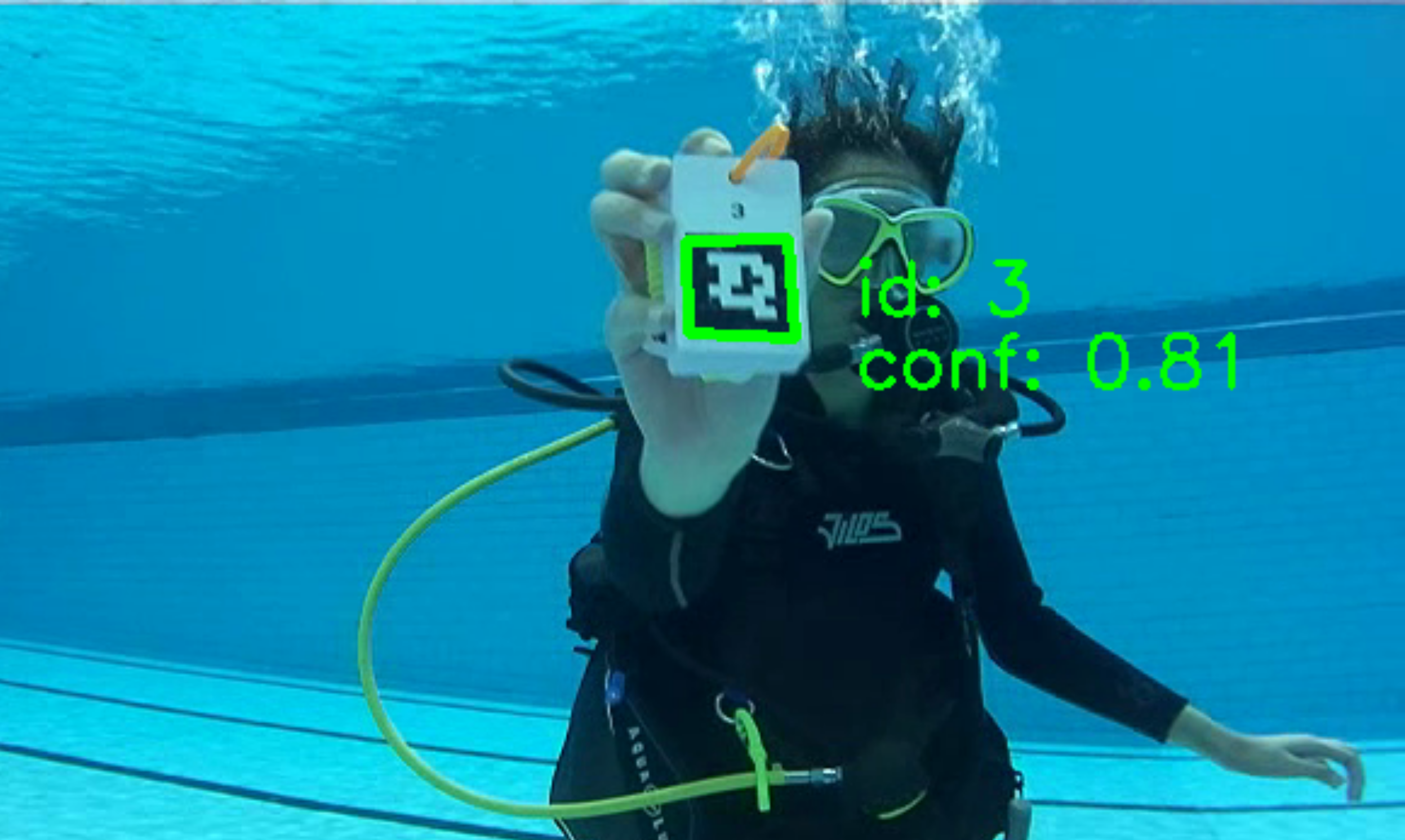}
        \caption{}
        \label{fig:tag}
    \end{subfigure}
    \caption{Hand gesture recognition and AR Tag detection running on MeCO in an open-water and closed-water environment respectively. Fig.~\ref{fig:gesture} shows the gesture is correctly recognized by our two stage recognition system. Fig.~\ref{fig:tag} shows the tag detection results using the artoolkitX tool.}
    \label{fig:gesture_and_tag_demo}
\end{figure}

\noindent \textbf{Gesture and AR Tag Control.} We tested \textit{in situ} mission reconfiguration by evaluating MeCO's onboard H2R communication modalities. MeCO utilizes two vision-based methods to receive commands from a diver underwater: \textit{Hand Gesture Recognition} and \textit{AR Tag Detection}. Both systems provide at least five different tokens for performing tasks, such as selecting items from a menu. Each token can be associated with a unique command for \meco. Fig.~\ref{fig:gesture_and_tag_demo} shows sample images from the H2R communication modality.

\noindent\textit{Hand Gesture Recognition.} The hand gesture recognition system includes two main components: hand detection and gesture classification. The region of the hand is first detected by YOLOv7-tiny \cite{yolov7}. Then, the detected regions are cropped from the images and classified by a multitasking network. For the multitasking network, we employ the Generalized Efficient Layer Aggregation Network (GELAN) proposed by YOLOv9 \cite{yolov9} as the backbone along with the transformer encoder to predict the gesture and its pose simultaneously. We trained the network jointly on the classification and pose estimation tasks, with the pose estimation task serving as an auxiliary task that is not used in the application. Features enriched by the pose estimation task can contribute to higher classification accuracy with fewer parameters and operations, making it suitable for deployment on edge devices. We trained the detector on data collected during our experiments, and the classifier on the HaGRID \cite{hagrid} large-scale dataset for hand gesture recognition.

\noindent \textit{AR Tag Detection.} We wrapped the artoolkitX tool \cite{artool} with our ROS 2 package to identify AR tags. By configuring different predefined tag patterns into our system using artoolkitX, we can obtain the identity, position, and pose of each tag in real-time from the camera data. This enables us to identify the message from a diver and execute corresponding commands. We can also execute more complex commands by combining hand gestures with tag detection, providing additional input modalities to our system.

\noindent \textbf{\meco Autonomous Vision (MAV)}. To test the ability of \meco to take an external stimulus and create autonomous motion, we conducted an experiment that had \meco follow a test target around bodies of water. An orange inflatable lift-bag was used as a target which \meco could see through its stereo camera. A color threshold was applied to detect the bag, creating a mask of the object. The pixel locations in the mask were cross-referenced with a generated depth map to determine a target vector, which was then fed to the controller, ensuring \meco did not approach closer than $0.5$~m. In the first test, \meco was tested in a local lake with severely degraded visual conditions, see Fig.~\ref{fig:mav_lake}. In this case \meco executed a number of turns, pitch maneuvers and speed changes to follow the target as it was moved through the water. Subsequently, the experiment was then repeated in the open ocean with different visual conditions, see Fig.~\ref{fig:mav_ocean} and again, followed the target as designed.

\begin{figure}[t]
    \centering
    \includegraphics[width=0.75\linewidth]{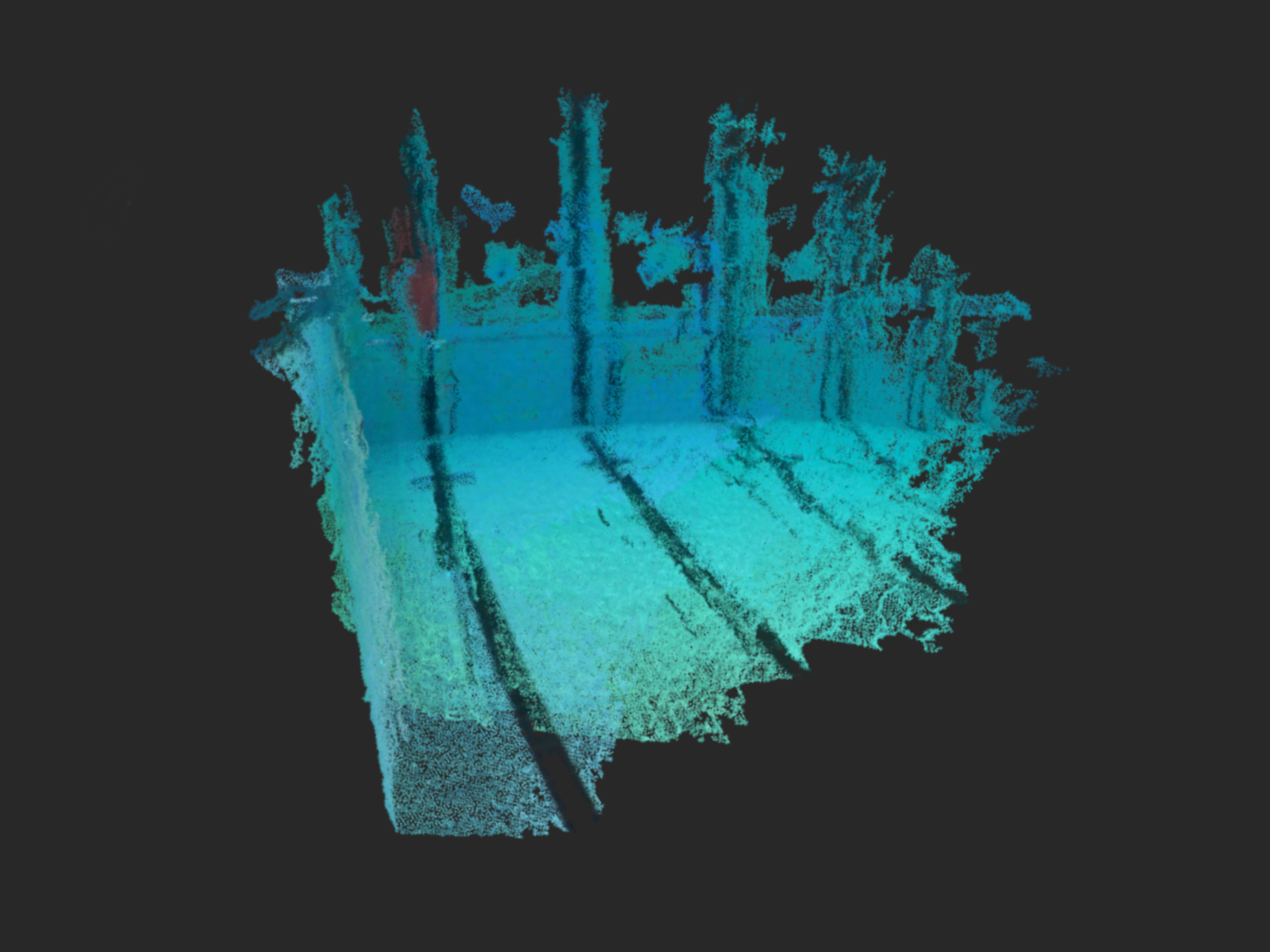}
    \caption{Point cloud map generated from Visual SLAM in a closed-water environment, demonstrating successful map generation in a low feature environment.}
    \label{slam_features}
\end{figure}

\noindent \textbf{Visual SLAM}. Visual SLAM (Simultaneous Localization and Mapping) has been successfully deployed on \meco, enabling real-time environmental mapping. This implementation utilizes RTAB-Map, an open-source ROS library for real-time appearance-based mapping \cite{rtabmap}. Visual-inertial odometry is performed through feature matching between keyframes captured by the stereo camera. Features are extracted using the Good Features to Track (GFFT) algorithm \cite{gftt}, while stereo correspondences between the left and right camera images are computed via optical flow using the iterative Lucas–Kanade method \cite{lucas}. IMU sensor data is incorporated to enforce gravity constraints, enhancing odometry accuracy. Loop closure detection is achieved using a bag-of-words approach \cite{bow}, improving map consistency and minimizing drift. 
This system successfully generated a map from ROS bag data recorded during a closed-water experiment, with the stereo camera detecting and integrating objects into the map from distances of up to three meters. This closed water environment lacked visual features for highly accurate tracking, yet the mapping was still successful (Fig.~\ref{slam_features}). 

\noindent \textbf{Trash Detection.} 
We deployed deep learning models on \meco to detect objects underwater in closed-water and open-water environments. We integrated the models with HREyes in the closed-water environment and SIREN in the open-water environment to provide real-time feedback. We picked different UHRI modes of feedback to demonstrate both audio and visual feedback capabilities of \meco.

\begin{figure}[t]
    \centering
    \begin{subfigure}{0.35\columnwidth}
        \centering
        \includegraphics[width=\textwidth]{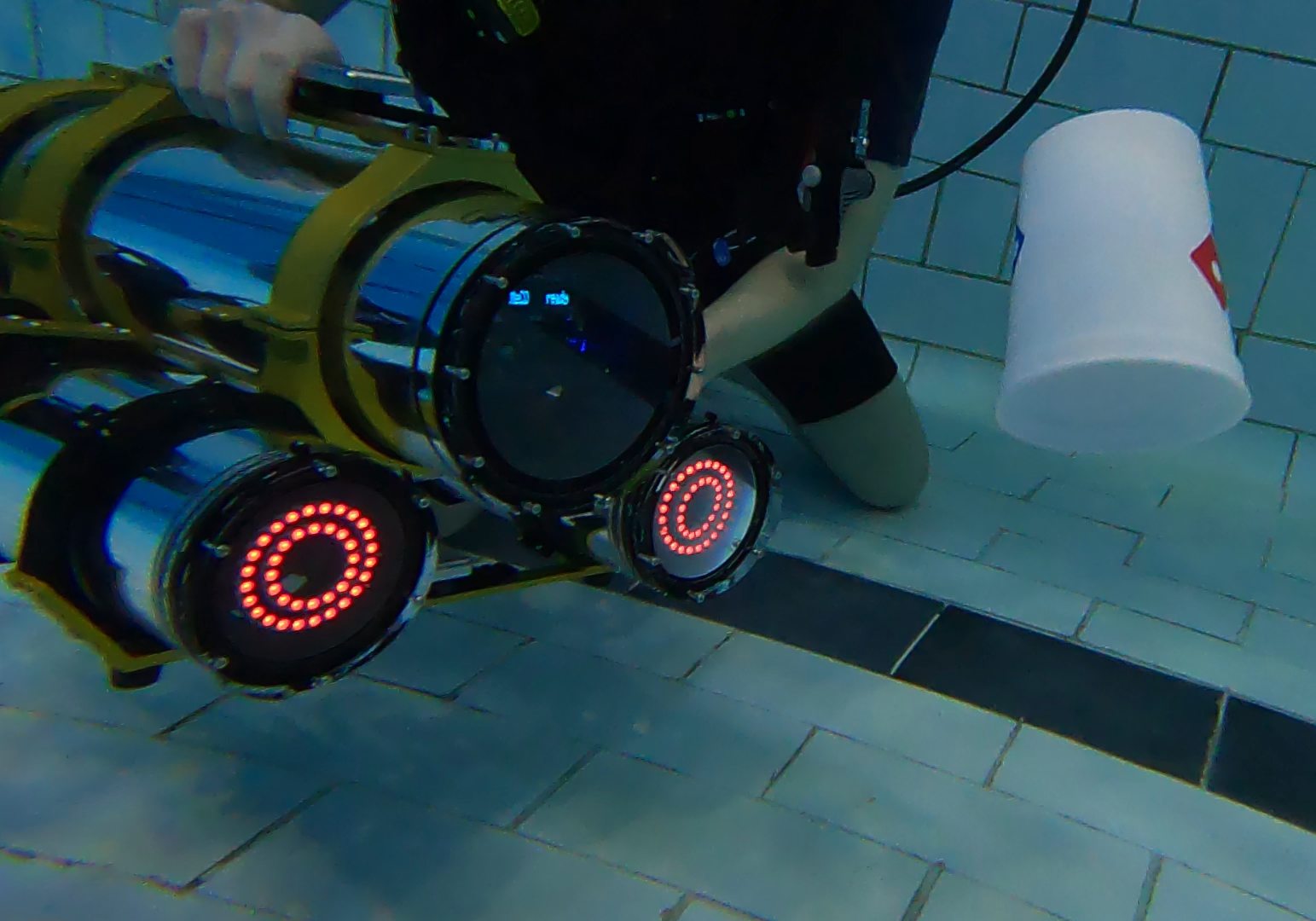}
        \caption{}
        \label{fig:trash_hreye}
    \end{subfigure}    
    \begin{subfigure}{0.35\columnwidth}
        \centering
        \includegraphics[width=\textwidth]{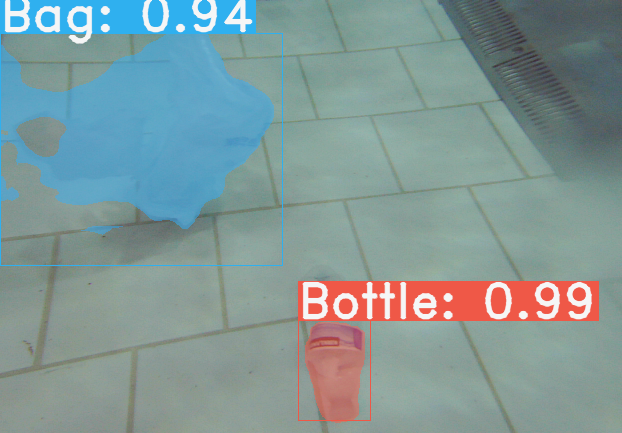}
        \caption{}
        \label{fig:vis_trash}
    \end{subfigure}
    \caption{Trash detection running on MeCO in a closed-water environment. Fig.~\ref{fig:trash_hreye} shows HREyes providing real-time detection feedback to the user. The LEDs display red after detecting a cup. Fig.~\ref{fig:vis_trash} is an example of a labeled frame captured using a separate underwater sensor system with an identical stereo camera.}
    \label{fig:trash_exp}
\end{figure}

\noindent \textit{Closed-water experiment}. We trained YOLACT \cite{bolya2019yolact} with a synthetic dataset created using the IBURD framework \cite{hong2025iburdimageblendingunderwater}. This dataset contains $7,982$ images across seven object classes. The objects in the dataset have unique LED colors associated with them to provide visual feedback: cup - red, mug - green, plastic bottle - blue, starfish - cyan, can - yellow, plastic bag - purple, glass bottle - white. Fig.~\ref{fig:trash_hreye} shows HREye feedback after detecting the object in real-time, and Fig.~\ref{fig:vis_trash} provides an example of the segmentation masks predicted using YOLACT.

\begin{figure}
\centering
\includegraphics[width = 0.6\linewidth]{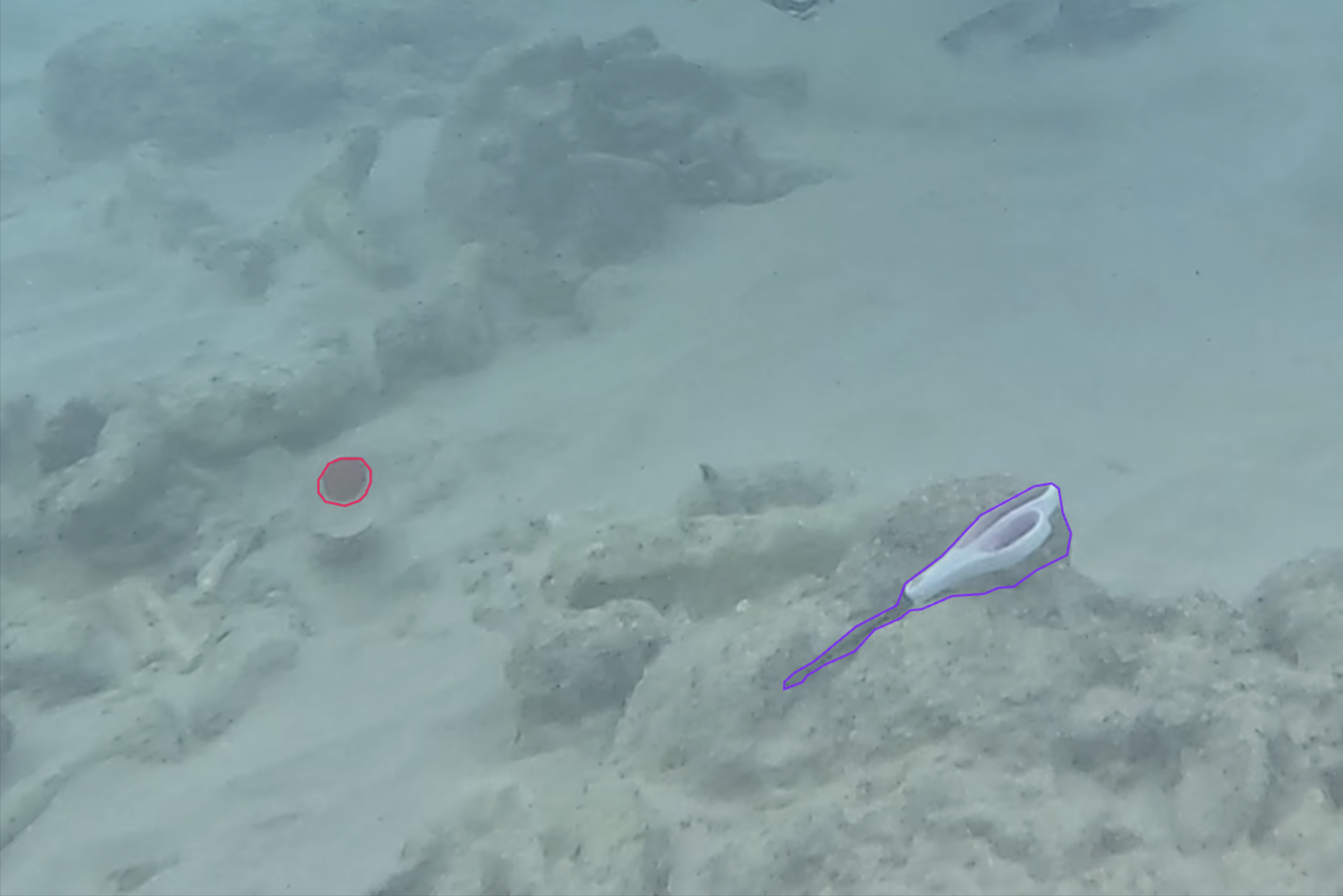}
\caption{COU instance segmentation running in an open-water environment. This image is an example of a predicted frame using COU, off the coast of Barbados. The objects are Scissors on the right and a Bottle on the left.}
\label{fig:vis_cou}
\end{figure}

\noindent \textit{Open-water experiment}. We used YOLOv9c \cite{yolov9} for underwater object detection in the Caribbean Sea off the coast of Barbados. We employed a different deep network compared to the closed-water experiments to demonstrate the flexibility of \meco in supporting various model architectures. 
This demonstrated the extensibility of the containerized software architecture by actively switching between deep networks which required different Python packages.
We fine-tuned the model using the common objects underwater (COU) dataset \cite{cou} and a subset of the COCO \cite{coco} dataset. The subset included classes that were relevant in the underwater domain \eg person.
Fig.~\ref{fig:vis_cou} shows an example of object segmentation on image from the open-water field experiment. Real-time feedback using the audio UHRI capability of SIREN is demonstrated in the attached video.

\noindent \textbf{End-to-end capability demonstration}. 
While the experiments mentioned above demonstrate the flexibility of the \meco platform in various individual use cases, we also combined these functionalities to demonstrate and develop the autonomous behavior of \meco. During the open-water experiments of MAV described above, multiple capabilities were combined to form a sophisticated mission that demonstrates the extensibility of \meco for complex UHRI research and applications. During these experiments, we performed the following:

\begin{itemize}
    \item Two divers swam with \meco to a pre-determined deployment point in the open-water environment for untethered operations
    \item One diver maintained \meco in place, while the other diver used a hand-gesture to start the MAV behavior 
    \item \meco followed the diver swimming with the orange target in the open sea
    \item The target diver disarmed the motors and deactivated the MAV protocol with the hand-gesture UHRI interface
    \item \meco's small, co-AUV form factor permitted the divers to swim \meco in a powered-off state to a different area and begin and autonomously complete a new mission using the hand-gesture system.
\end{itemize}
These end-to-end mission experiments demonstrated the ability to use and expand the functionality of \meco for real-world experiments.

\section{Conclusion}

This manuscript presents the reconfigurable hardware and electrical design of MeCO, the vehicle’s movement capabilities, the software architecture, the Unity simulation, the system’s UHRI methods, and the experiments implemented on the platform. The development of \meco advances AUV research and provides another option for an affordable open-source AUV. The maneuverability, upgraded computing system, modular design, and UHRI features make \meco valuable for various underwater research applications that necessitate collaboration with divers. Planned future improvements include installation of wide-baseline stereo cameras, 5G radio for surface connectivity, and waypoint navigation, among others.

\bibliographystyle{ieeetr}
\bibliography{bibliography}
\end{document}